\documentclass[12pt]{article}
\usepackage{epsfig}

\begin{document}

\title{Language identification of controlled systems: Modelling, control and
anomaly detection}
\author{J. F. Martins\thanks{%
Laborat\'{o}rio de Mecatr\'{o}nica, Instituto Superior T\'{e}cnico, Av.
Rovisco Pais, 1096 Lisboa Codex, Portugal} \thanks{%
Escola Superior de Tecnologia, Instituto Polit\'{e}cnico de Set\'{u}bal, R.
do Vale de Chaves, Estefanilha, 2910 Set\'{u}bal, Portugal}, J. A. Dente%
\footnotemark[1]  , A.J. Pires\footnotemark[2]  \footnotemark[1]  and R.
Vilela Mendes\footnotemark[1] }
\date{}
\maketitle

\begin{abstract}
Formal language techniques have been used in the past to study autonomous
dynamical systems. However, for controlled systems, new features are needed
to distinguish between information generated by the system and input
control. We show how the modelling framework for controlled dynamical
systems leads naturally to a formulation in terms of context-dependent
grammars. A learning algorithm is proposed for on-line generation of the
grammar productions, this formulation being then used for modelling, control
and anomaly detection. Practical applications are described for
electromechanical drives. Grammatical interpolation techniques yield
accurate results and the pattern detection capabilities of the
language-based formulation makes it a promising technique for the early
detection of anomalies or faulty behaviour.
\end{abstract}

\section{Introduction}

Formal language theory has been used in the past to study autonomous
systems, the grammatical complexity of the sequences generated according to
some coding being used to characterize the complexity of the dynamical
system. References \cite{Wolfram} to \cite{Moore} are a representative
sample of some of this work. However, most systems used in technological
applications are not autonomous, but controlled dynamical systems. Therefore
one has to distinguish, from the start, between the information that is
generated by the dynamics of the system and the one that depends on the free
will of the controlling operator. This has some implications on the nature
of the formal entities that are used to code the system variables and on the
nature of the languages that are generated. Another important requirement
for practical applications is the possibility to extract the grammars by on
line learning from the data generated by the system, independently of
whether or not an analytic model is known.

In Section 2, after a short introduction to some formal language concepts,
we show how the modelling framework for controlled dynamical systems leads,
almost uniquely, to a context-dependent grammatical formulation. A learning
algorithm is then proposed for on-line generation of the productions of the
grammar. The following sections illustrate how this formulation may be used
for modelling, control and anomaly detection.

Most of our applications concern electromechanical drives. Using grammatical
interpolation, as described in Section 3, one may obtain quite accurate
results, even with small grammars. However the main capabilities of a
language-based formulation lies in the pattern detection domain. Therefore
early detection of anomalies or faulty behaviour is probably the most
promising domain of application for this technique.

\section{Grammars for controlled dynamical systems}

\subsection{Formal language concepts}

Many fundamental contributions to the formal language field originate from
the work of Chomsky \cite{Chomsky1} - \cite{Rozenberg}, whose theory of
formal grammars had a major influence in the development of the subject. 
\textit{Grammatical inference}, that is, the development of algorithms which
extract grammatical information from examples, is a concept that goes back
to Gold's work \cite{Gold}. Since then a great deal of work has been done,
which can be found in several surveys \cite{Angluin} - \cite{Sakakibara}

To apply grammatical inference procedures, a dynamical system must be
considered as an entity (linguistic source) capable of generating a specific
language. The \textit{grammar} $G$ of the language is the set of rules that
specifies all the \textit{words} in the language and their relationships.
Once the grammar is found, the grammar itself is a \textit{model} for the
source. To define a grammar $G$ one specifies a \textit{terminal alphabet},
a \textit{non-terminal alphabet}, a \textit{start symbol} and a \textit{set
of productions} 
\begin{equation}
G=\left\{ \Sigma _{T},\Sigma _{N},S,P\right\}  \label{2.1}
\end{equation}

\textit{Terminal alphabet (}$\Sigma _{T}$\textit{)}: set of symbols that
make up the \textit{words}, a word being a string of symbols.

\textit{Non-terminal alphabet (}$\Sigma _{N}$\textit{)}: set of auxiliary
symbols that are used to generate the words by the production rules.

\textit{Start symbol} ($S$): a special non-terminal symbol used to start the
generation of words.

\textit{Productions (}$P$\textit{)}: set of substitution rules (denoted $%
a\rightarrow b$) used to generate the allowed words in the language.

Example: the grammar 
\begin{equation}
\left\{ \Sigma _{T}=\left( a,b\right) ,\Sigma _{N}=\left( A\right)
,S,P=\left( S\rightarrow bA,A\rightarrow aA,A\rightarrow a\right) \right\}
\label{2.2}
\end{equation}
generates the language with words consisting of a $b$ symbol followed by any
number of $a$ symbols.

The set of productions encodes the dynamics of the system that generates the
language. Any word that can be derived from the start symbol by a sequence
of productions of the grammar is said to be in the language generated by the
dynamical system.

\textit{Grammatical inference} is an algorithm by which a grammar is
inferred from a set of sample words produced by the dynamical system
considered as the linguistic source. Notice that there is not a unique
relation between a language and the grammar, because distinct grammars may
generate the same language.

Grammatical inference, in general, is the identification of a grammar from a
set of positive and negative examples. One may also consider a scheme where
a ''teacher'' answers questions concerning the language to be inferred.
However, in dynamical system identification by grammatical inference, we are
in the restricted setting of language identification from positive examples
alone. Therefore a finite sample cannot characterize an infinite language.
It may at most specify the class of languages, which possesses that finite
sample as an allowed one.

A practical problem, when learning from a finite set of examples, is the
fact that some relevant productions may never be inferred. If a metric is
defined in the space of words, this problem may be solved to some extent by 
\textit{grammatical interpolation techniques}. Another practical problem
arises when the system under study is perturbed by noise. Then, productions
may be inferred which are not characteristic of the system, but are a result
of the noise perturbation. The solution used in our applications is to keep
as valid only those productions that appear many times in the learning
process. The threshold for acceptance of a production depends of course on
the size of the training set and the estimated noise intensity.

We work here in the \textit{phrase-structure grammar} setting as defined by
Chomsky. Notice however that more general systems like \textit{programmed
grammars}, \textit{cooperating distributed grammars} or \textit{contextual
grammars} \cite{Rozenberg} may also find useful applications in the
characterization of dynamical systems. In particular, programmed languages
and the interpretation of controls as labels, could be a viable alternative
to our identification of the control variables with non-terminal symbols.

\subsection{Grammatical inference in controlled dynamical systems}

A model for a controlled dynamical system has the general form 
\begin{equation}
\begin{array}{lll}
\stackrel{\bullet }{x} & = & f\left( x,U\right) \\ 
y & = & h\left( x\right)
\end{array}
\label{2.1a}
\end{equation}
or, considering the time divided into discrete steps 
\begin{equation}
\begin{array}{rll}
x_{k+1} & = & f\left( x_{k},U_{k}\right) \\ 
y_{k} & = & h\left( x_{k}\right)
\end{array}
\label{2.1b}
\end{equation}
where $x$ is the state variable, $y$ the output (or observed) variable and $%
U $ the input (or control) variable. Eqs.(\ref{2.1a} - \ref{2.1b}) also
establish a functional relationship between the output variables at
different times 
\begin{equation}
\begin{array}{rll}
y_{k+1} & = & g\left( y_{k},U_{k}\right)
\end{array}
\label{2.1c}
\end{equation}
However, in most systems used in technology, not all state variables are
observable. Therefore Eq.(\ref{2.1c}) does not provide a complete
specification of the system. In general, specification of the dynamics in
terms of the output variables requires a set of functional relationships
involving many time steps in the past, namely 
\begin{equation}
\begin{array}{lll}
y_{k+1} & = & g_{0}\left( U_{k}\right) \\ 
y_{k+1} & = & g_{1}\left( y_{k},U_{k}\right) \\ 
y_{k+1} & = & g_{2}\left( y_{k},y_{k-1},U_{k}\right) \\ 
y_{k+1} & = & g_{3}\left( y_{k},y_{k-1},y_{k-2},U_{k}\right) \\ 
& \vdots & 
\end{array}
\label{2.1d}
\end{equation}
It is this structure, which is required by dynamical considerations on
actual controlled systems, that leads in a natural way to our proposal of $%
p- $type productions.

To develop a grammatical description and a grammatical inference algorithm
for controlled dynamical systems three steps are required. First, the
quantification of the variables, then the specification of the nature of the
productions and finally a learning algorithm to extract the productions from
the experimental data.

\begin{itemize}
\item  \textit{Quantification}
\end{itemize}

Quantification refers to the creation of alphabets for the output variable $%
y $ and the control variable $U$. In our approach we associate the terminal
alphabet $\Sigma _{T}$ to the output variable $y$ and the non-terminal
alphabet $\Sigma _{N}$ to the control variable $U$. Let $n$ be the number of
terminal symbols and $m$ the number of non-terminal symbols. A
quantification of the variables is made, in a discrete way, dividing the
variables range in equal intervals and associating each interval to a symbol
in the alphabet 
\begin{equation}
\begin{array}{lllll}
y & \leftrightarrow & y_{i}\in \Sigma _{T} &  & i=1\cdots n \\ 
U & \leftrightarrow & U_{j}\in \Sigma _{N} &  & j=1\cdots m
\end{array}
\label{2.4}
\end{equation}

\begin{itemize}
\item  \textit{Productions}
\end{itemize}

$p-$\textit{type productions} are defined to be substitution rules of the
form 
\begin{equation}
y_{1}\cdots y_{p}U_{k}\longrightarrow y_{1}\cdots y_{p}y_{p+1}\delta
\label{2.5}
\end{equation}
where $y_{1}\cdots y_{p}$ is a sequence of terminal symbols, $U_{k}$ a non
terminal symbol and $\delta $ a special non terminal. $\delta $ is used to
allow the conclusion, or not, of a generated word, by the use of following
special set of productions 
\begin{equation}
\begin{tabular}{ccccc}
$\delta $ & $\longrightarrow $ & $U_{j}$ & $,$ & $j=1\cdots m$ \\ 
$\delta $ & $\longrightarrow $ & $\lambda $ &  & 
\end{tabular}
\label{2.6}
\end{equation}
where $\lambda $ denotes the empty symbol.

A $p-$type production codes the evolution of the output variable, depending
on its $p$ past values and on the value of the control variable $U$. There
is therefore a functional relationships between the dynamics of the system
and the $p-$type productions.

\begin{itemize}
\item  \textit{Learning algorithm}
\end{itemize}

To obtain a sample of the language, a sequence of control signals is applied
to the system is such a way that the output variable $y$ takes values in a
sufficiently wide region. The signal evolution is then quantified as
described above and the following learning algorithm is used:

1. A $0-$type production is assumed for every newly occurring control symbol.

2. A new $(n+1)-$type production is generated each time the data conflicts
with the previously established $n-$type productions. The conflicting $n-$%
type productions are also promoted to $\left( n+1\right) -$type productions
or are deleted if there is not sufficient information on the past to do so.

To be able to revise the conflicting $n-$type productions, promoting them to 
$\left( n+1\right) -$type productions, we need to keep in a \textit{memory
window} the record of a certain numbers of past steps in the system
evolution. The length of the memory window need not be larger than $p_{\max
} $ ($p_{\max }$ being the highest order allowed for the productions). In
general there is not great harm in having zero-length for the memory window
and just deleting all the conflicting lower order productions. This is
because all the most relevant productions will reappear some time again in
the future.

As an example, consider the following symbol sequence obtained from a data
sample 
\begin{equation}
\begin{array}{lllllllll}
U & \textnormal{variable} & : & A & B & A & A & B & A \\ 
y & \textnormal{variable} & : & e & d & c & b & d & e
\end{array}
\label{2.7}
\end{equation}
At time zero the algorithm analyzes the leading symbols of both the control
and the output variable. Since no other information is yet available, a $0-$%
type production $A\rightarrow e\delta $ is assumed. After analyzing the
second symbol, the algorithm establishes another $0-$type production $%
B\rightarrow d\delta $. The third symbol would yield a $0-$type production $%
A\rightarrow c\delta $. However, this production contradicts the previously
established $A\rightarrow e\delta $, because the grammar is not stochastic.
Therefore, in this case, a $1-$type production $dA\rightarrow dc\delta $ is
obtained. The $0-$type production $A\rightarrow e\delta $ is deleted because
no information is available on the past of the first symbol ($e$). At the
next step we obtain the production. $cA\rightarrow cb\delta $. Another
conflict arises when we reach the last symbol in (\ref{2.7}). Then a $2-$%
type production $bdA\rightarrow bde\delta $ is obtained and $dA\rightarrow
dc\delta $ is revised to $edA\rightarrow edc\delta $. Finally one is left
with 
\begin{equation}
\begin{array}{rll}
B & \rightarrow & d\delta \\ 
cA & \rightarrow & cb\delta \\ 
bdA & \rightarrow & bde\delta \\ 
edA & \rightarrow & edc\delta
\end{array}
\label{2.8}
\end{equation}
The learning algorithm is described by the flow chart in Fig.1.
\begin{figure}[tbh]
\begin{center}
\psfig{figure=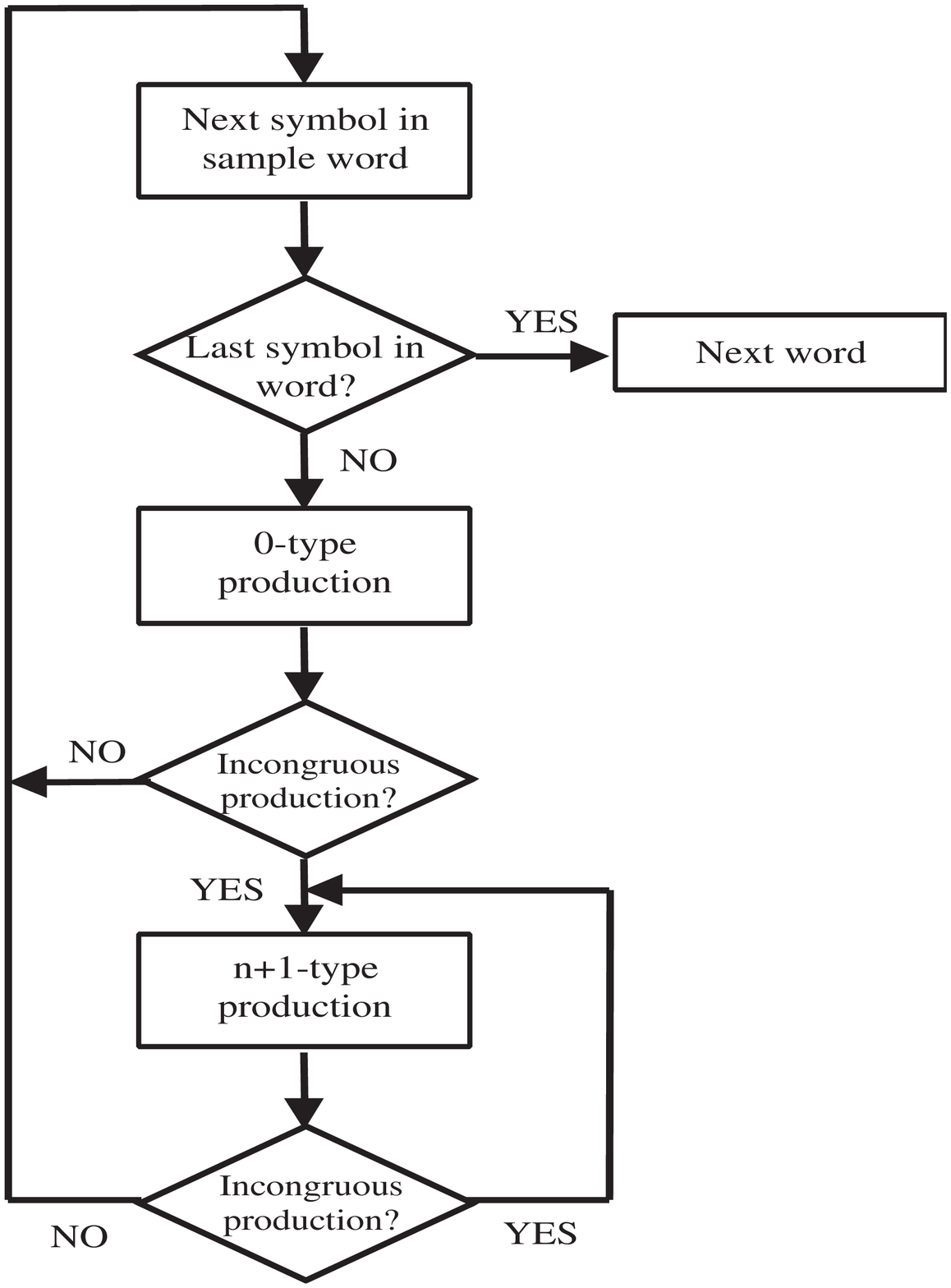,width=9truecm}
\end{center}
\caption[]{Flow chart of the grammatical inference learning algorithm}
\end{figure}
Unlike, for example, the rules for fuzzy logic inference, the structure of
the formal language productions, obtained by our algorithm, is not
established in advance. A mixture of different types of productions may be
obtained in the grammar. This feature provides flexible adaptation in
dynamical systems displaying distinct behavior in different regions of the
working space.

In a controlled system the two types of variables, output variables and
control variables, have a distinct nature. This is represented in our
formalism by the assignment of terminal symbols to the dynamical variables
and non-terminal symbols to the control. In turn, the productions represent
the action of the control in the context of the past and present dynamics of
the system. The languages generated in this way fall in the category of
context-sensitive languages\cite{Chomsky2}.

In practical applications the grammatical inference algorithm is completed
with \textit{grammatical interpolation} and a \textit{noise rejection}
mechanism. Grammatical interpolation is discussed in Section 3 and noise
rejection is simply implemented by keeping only those productions that
appear a sufficient number of times in the data. Also, for practical
purposes, an upper bound is put on the order of the productions that the
algorithm generates.

As a simple example, of how this methodology is used to model the dynamics,
we consider the following piecewise linear system

\begin{center}
\begin{equation}
y_{k+1}=y_{k}-y_{k-1}+2U_{k}\textnormal{ with }\left\{ 
\begin{array}{c}
y_{k}\in \left\{ 1,2,3,4\right\} \\ 
U_{k}\in \left\{ -1,0,1\right\}
\end{array}
\right.  \label{2.9}
\end{equation}
\end{center}

The working domain of this dynamical systems is spanned by three distinct
planes, depending on the values of the $U_{k}$ variable (Fig.2).
\begin{figure}[tbh]
\begin{center}
\psfig{figure=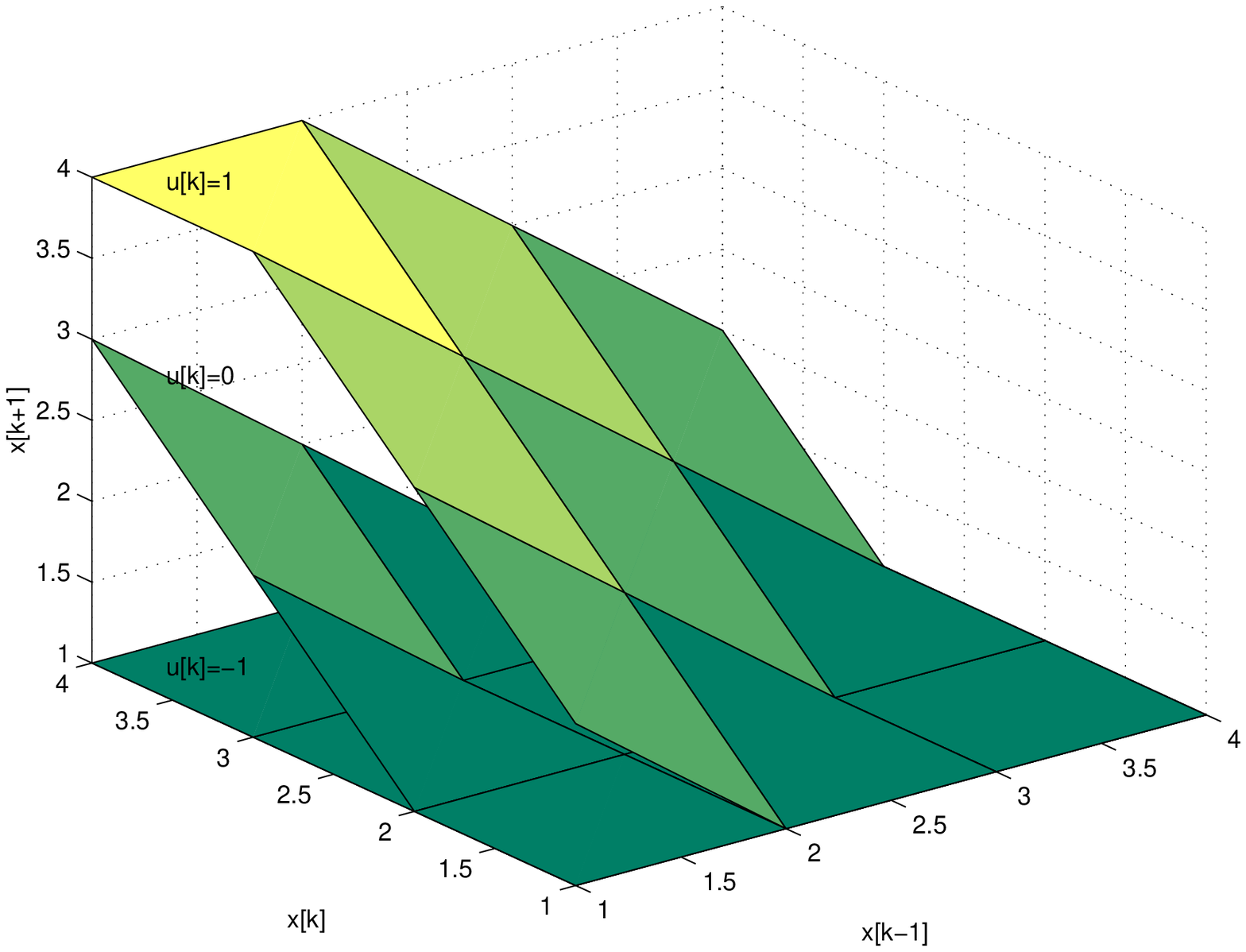,width=9truecm}
\end{center}
\caption[]{Working domain of a piecewise linear dynamical system}
\end{figure}
The assumed codification is

\begin{center}
\begin{tabular}{|c|c|}
\hline
$y_{k}\leftrightarrow \Sigma _{T}$ & $U_{k}\leftrightarrow \Sigma _{N}$ \\ 
\hline
1 $\leftrightarrow $ a & $-$1 $\leftrightarrow $\ A \\ \hline
2 $\leftrightarrow $\ b & 0 $\leftrightarrow $\ B \\ \hline
3 $\leftrightarrow $\ c & 1 $\leftrightarrow $\ C \\ \hline
4 $\leftrightarrow $\ d &  \\ \hline
\end{tabular}
\end{center}

The language generated by this dynamical system is represented by a grammar $%
G_{1}$, whose rewriting system has one $0-$type production, two $1-$type
productions and twenty four $2-$type productions.

This system was used to test our on-line learning algorithm with a random
input control signal $U_{k}$. Fig.3 shows the evolution of the productions,
extracted from randomly generated data, for the first 30 data points. One
sees, for example, at the data point 19, the cancellation of a $0-$type
production forced by a new $1-$type production. Also at data point 62 two
contradictory $2-$type productions are discarded.
\begin{figure}[tbh]
\begin{center}
\psfig{figure=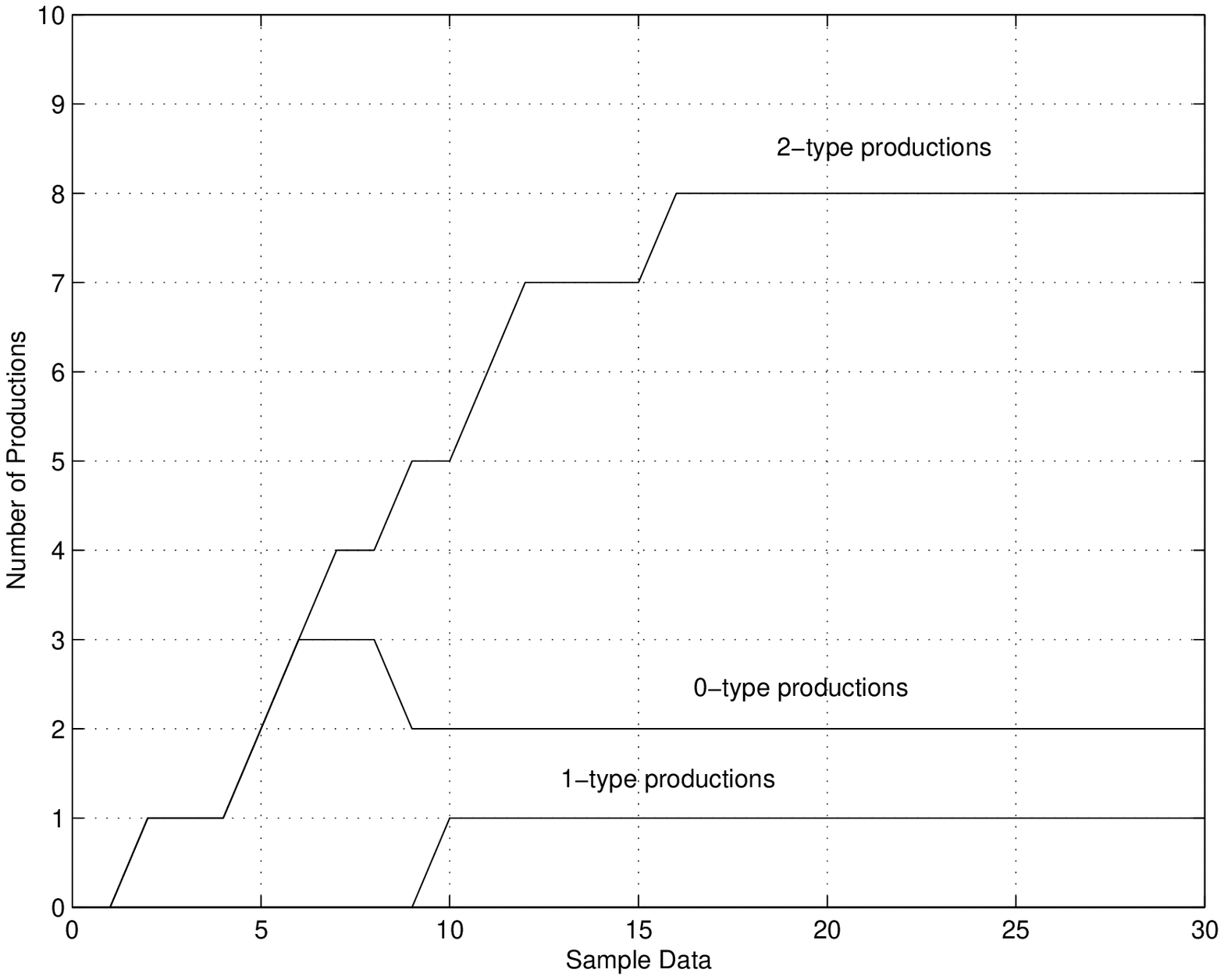,width=9truecm}
\end{center}
\caption[]{Evolution of the learned grammar productions for a piecewise linear
dynamical system}
\end{figure}

\section{Modelling an electrical drive}

The linguistic coding described before has been used, in our laboratory, to
model an experimental drive system, consisting of an electronically fed
induction machine.

The experimental system is depicted in Fig.4. It is composed of an induction
motor driven by a power inverter. Drive systems of this type are modelled
using the electromechanical power conversion theory \cite{Fitzgerald}. A
12-equations model represents the electrical drive system. The large number
of variables hinders any attempt to perform on-line learning without huge
computational costs. Therefore, one needs to simplify the electrical drive
model.
\begin{figure}[tbh]
\begin{center}
\psfig{figure=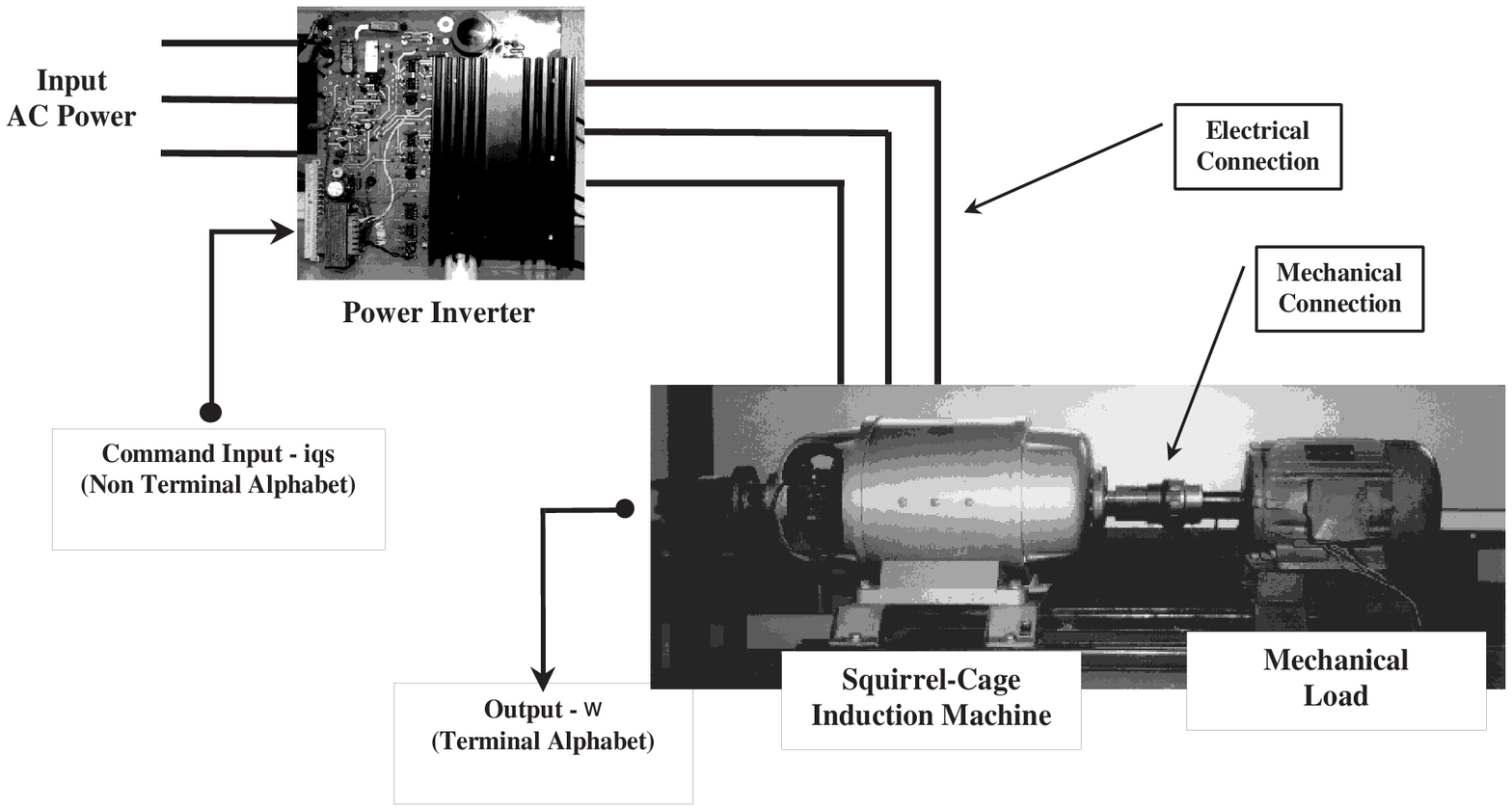,width=9truecm}
\end{center}
\caption[]{Electrical drive system}
\end{figure}

Considering an input variable internal control loop \cite{Martins1}, stator
currents may be assumed to be controlled. The model is then reduced from a
12th to a 3rd order system \cite{Martins2} which, in the rotor flux
reference frame, is 
\begin{equation}
\begin{array}{lll}
\stackrel{\bullet }{\psi }_{dr} & = & -\frac{1}{\tau _{r}}\psi _{dr}+\left(
\omega _{R}-\omega \right) \psi _{qr}+\frac{M}{\tau _{r}}i_{ds} \\ 
\stackrel{\bullet }{\psi }_{qr} & = & -\left( \omega _{R}-\omega \right)
\psi _{dr}-\frac{1}{\tau _{r}}\psi _{qr}+\frac{M}{\tau _{r}}i_{qs} \\ 
\stackrel{\bullet }{\omega } & = & \frac{M}{JL_{r}}\left( i_{qs}\psi
_{dr}-i_{ds}\psi _{qr}\right) -\frac{B}{J}\omega -T_{ext}
\end{array}
\label{3.1}
\end{equation}

The rotor fluxes ($\psi _{dr}$,$\psi _{qr}$) and the rotor speed ($\omega $)
are the state variables, and the stator current components ($i_{ds}$,$i_{qs}$%
) the control variables. $\tau _{r}$ is the rotor time constant, $M$ the
mutual (stator-rotor) induction coefficient, $J$ the inertia coefficient, $%
L_{r}$ the rotor self-inductance coefficient, $B$ the friction coefficient
and $T_{ext}$ the load torque. With the reference frame rotor flux given by 
\begin{equation}
\omega _{R}=\omega +\frac{M}{\tau _{r}}\frac{i_{qs}}{\psi _{r}}  \label{3.2}
\end{equation}
one obtains from (\ref{3.1}) 
\[
\begin{array}{lll}
\stackrel{\bullet }{\psi }_{qr} & = & -\frac{1}{\tau _{r}}\psi _{r}+\frac{M}{%
\tau _{r}}i_{ds} \\ 
\stackrel{\bullet }{\omega } & = & \frac{M}{JL_{r}}i_{qs}\psi _{r}-\frac{B}{J%
}\omega -T_{ext}
\end{array}
\]

To obtain relevant training sets, two types of information, qualitative and
quantitative, should be acquired. Qualitative information is obtained from
the mathematical models. This leads to the choice of the relevant variables
that describe the system behavior. This choice is very important to insure
the existence of a functional relationship representative of the drive
behavior\cite{Martins3}. Otherwise both learning and recognition would be
impossible.

To obtain quantitative information experimental data is acquired. To obtain
the data an excitation signal must be chosen. A possible approach would be
to use a pseudo-random binary signal\cite{Teck}. However, this signal is not
the best choice for drive systems because it is filtered by mechanical time
constants. It is better to use sinusoidal signals of different amplitudes
and frequencies. In this way, frequencies and amplitudes may be adjusted
within the limits of the drive response, thus avoiding the filtering problem
and collecting data that adequately spans the operating domain.

A grammar of the drive language is inferred from the (control) input -
output experimental information. The $i_{ds}$ current is kept constant, a
non terminal alphabet is established from the quantification of the $i_{qs}$
current and a terminal alphabet established from the output speed signal ($%
\omega $). A representative training set is obtained with a sinusoidal
reference signal containing a combination of different amplitudes and
frequencies. Fig.5 shows the evolution, in the input/output space, of the
training and test data sets, the test data being displayed in bold.

The acquisition of the training set is an essential step in obtaining a good
knowledge of the system to be modelled. The training set must cover a
representative part of the entire working domain.
\begin{figure}[tbh]
\begin{center}
\psfig{figure=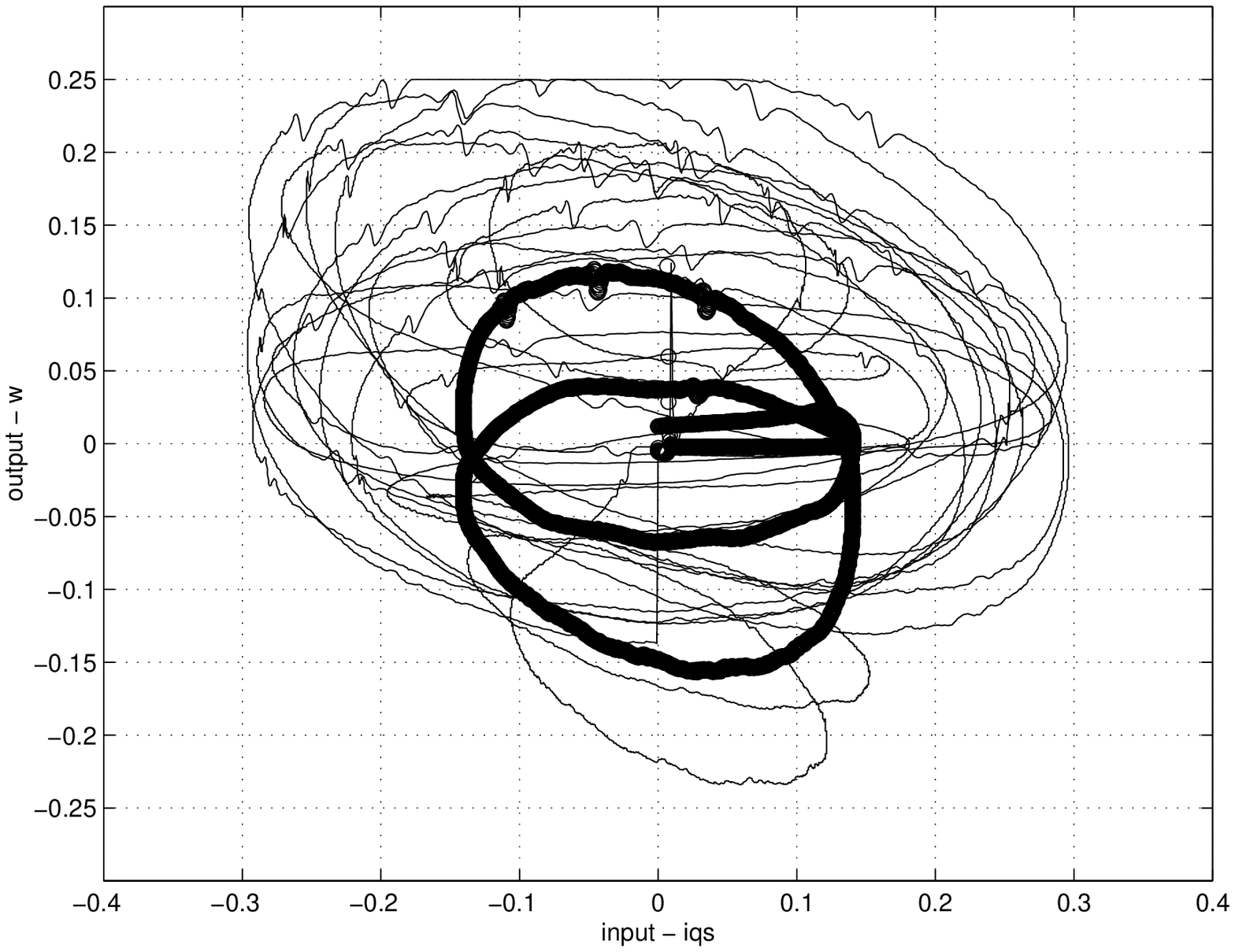,width=9truecm}
\end{center}
\caption[]{Training and test data set for the electrical drive}
\end{figure}

\subsection{Recognition results}

To test the validity of the recognition process, two alphabets were
considered. A quantification is established for both input and output
variables, yielding a 60 symbol alphabet.

The inferred grammar contains 183 productions distributed in the following
way: 5 $0-$type productions, 156 $1-$type productions and 22 $2-$type
productions. To avoid the influence of noise on the inferred productions, a
production is considered valid only if it appears more than a certain number
of times, this parameter depending on the size the training set. This
procedure works as a filter on noise perturbations.

Applying this grammar to the test set, we obtain the recognition results
shown in Fig.6. The actual electromechanical drive speed is shown as a
dotted line and the grammar response as a continuous one.
\begin{figure}[tbh]
\begin{center}
\psfig{figure=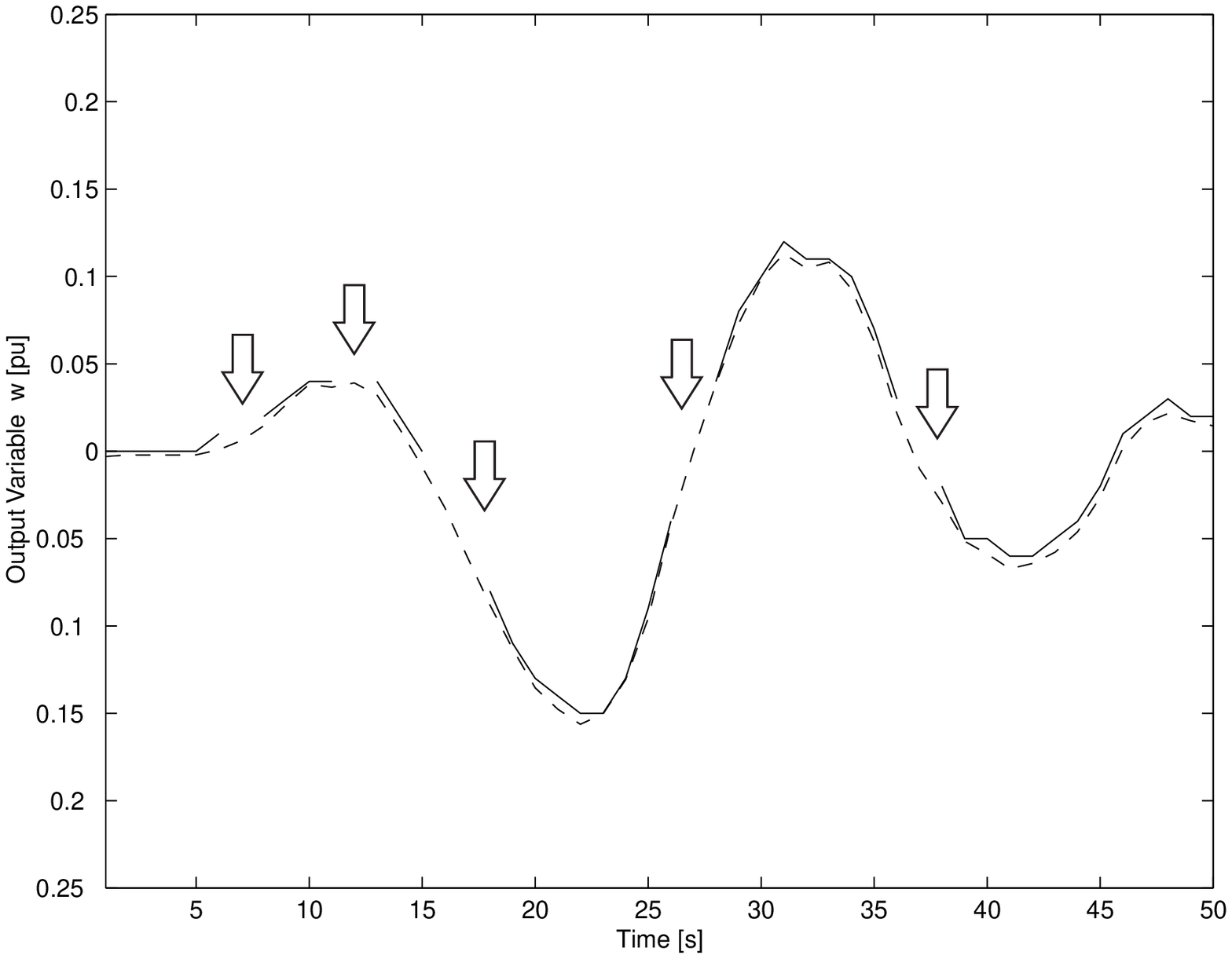,width=9truecm}
\end{center}
\caption[]{Formal language recognition (60-letter alphabet)}
\end{figure}

Apart from the quantification error, the recognition results may be
considered satisfactory, because the evolution of the drive speed and the
grammar results are similar. However, we must point out that there are some
situations, indicated by the white arrows in Fig.6, where the grammar does
not provide any answer. This happens when no production was inferred
representing that particular input/output relationship. As explained later,
a generalization method is used to establish non-learned productions.

For the second alphabet, a quantification interval ten times higher is
assumed for both input and output variables, yielding now a 6-symbol
alphabet, instead of the previous 60 symbols. The inferred grammar contains
36 productions, distributed in the following way: 0 $0-$type productions, 4 $%
1-$type productions and 32 $2-$type productions. Applying this grammar to
the test set the recognition results are those shown in Fig.7. As before,
the electromechanical drive speed is shown as a dotted line and the grammar
response as a continuous one. The white arrows denote the absence of
applicable productions.
\begin{figure}[tbh]
\begin{center}
\psfig{figure=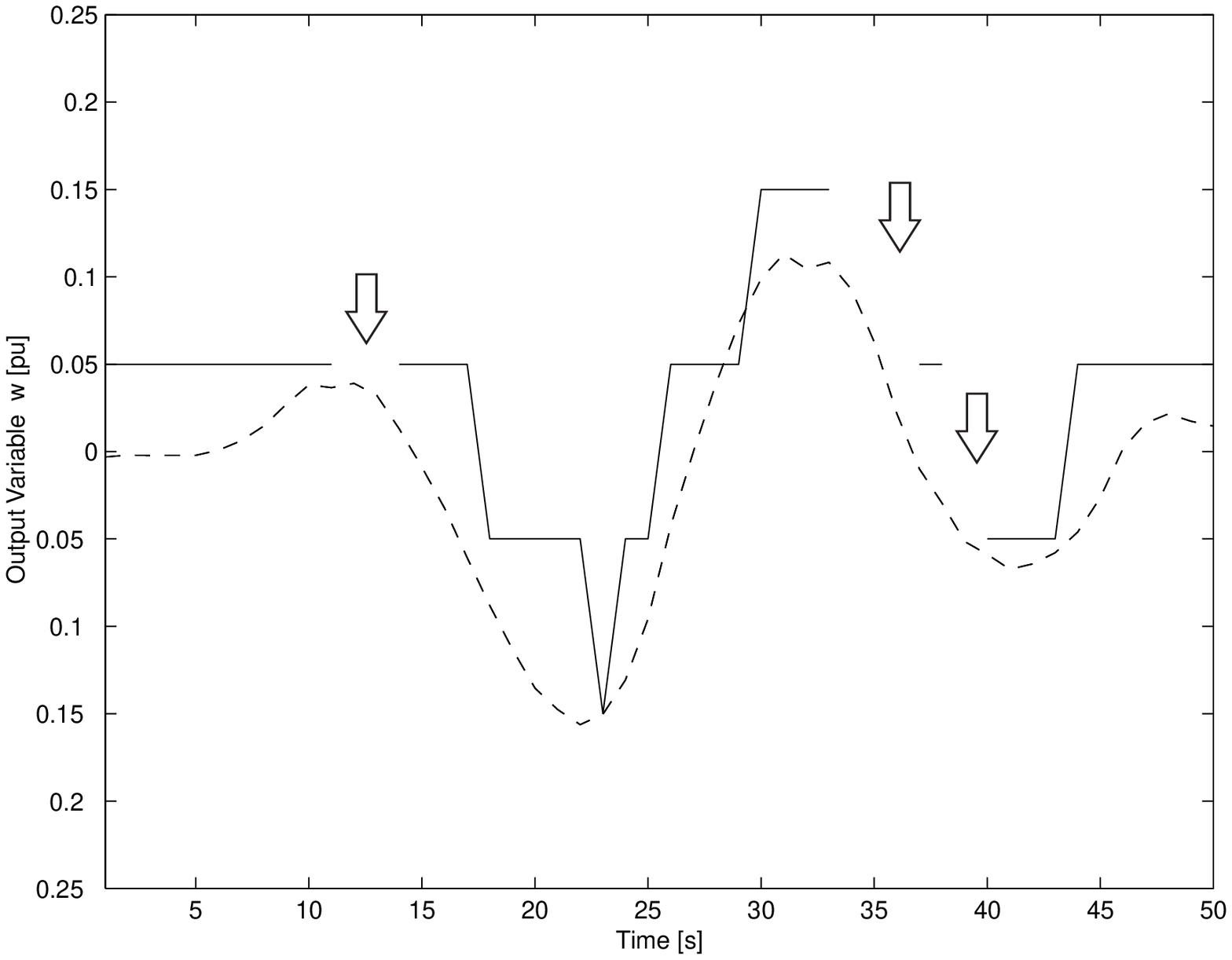,width=9truecm}
\end{center}
\caption[]{Formal language recognition (6-letter alphabet)}
\end{figure}

As expected, the quantitative modelling performance deteriorates, there
being a larger difference between the measured drive speed and the
grammatical reconstruction. However, if the important feature is the
qualitative recognition of the drive speed, the results may still be
considered as satisfactory. For a qualitative recognition process a limited
alphabet is not a serious drawback. However, as in the larger grammar, the
problem of non-existent productions is present here.

The discrete quantification performed within the grammatical inference
algorithm implies a discrete response when recognition of the drive is
performed. In order to improve the quantitative recognition capabilities of
the algorithm two solutions may be used. The first is to increase the number
of symbols in the alphabet to make a more accurate quantification.
Alternatively one may use numerical interpolation of the productions. The
first solution increases the number of productions thus slowing the
recognition process. The second solution does not increase the number of
productions.

To illustrate the operation of the interpolation process consider again the
60-symbol alphabet and another training set which is half the size of the
one used for the results in Fig.6. After inferring the correspondent
grammar, the recognition results are shown in Fig.8. It is seen that the
recognition process has considerable faults because, for the reduced
training set, the number of productions that is obtained is considerably
smaller. A smaller training set led to a faulty grammar, which fails to
recognize some words. However because the symbols in our alphabets are in
correspondence with numerical values of the variables domain, a metric may
be defined in the space of words. New productions may then be obtained, from
those inferred from the smaller sample, by interpolation.
\begin{figure}[tbh]
\begin{center}
\psfig{figure=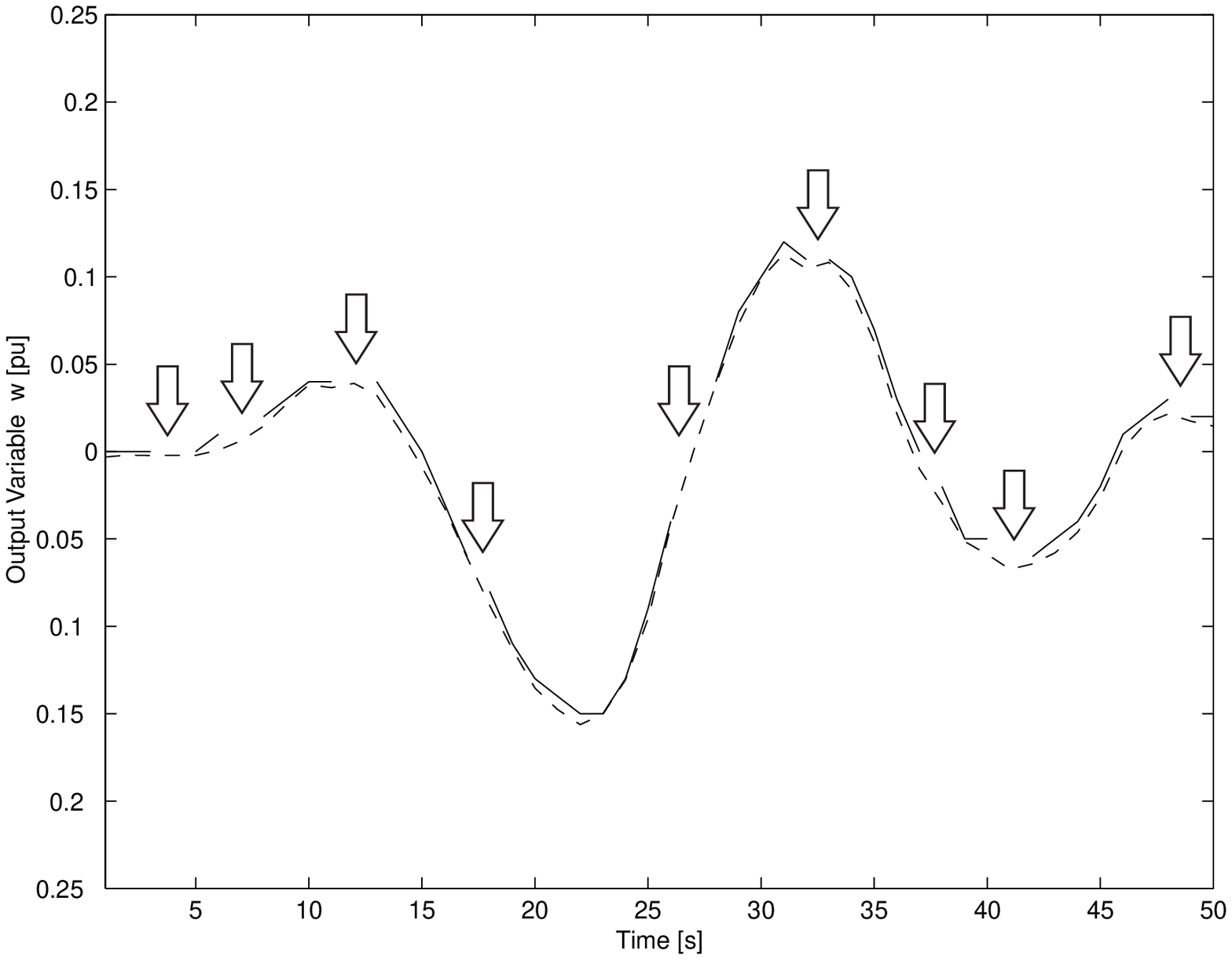,width=9truecm}
\end{center}
\caption[]{Formal language recognition with a poor training set}
\end{figure}

\textit{Grammatical interpolation} establishes new productions by a
structural matching procedure. The main idea of structural matching is based
on a measure of similarity between the unknown input pattern and the
available data structures. The measure of similarity is the distance between
the non-existent production and the nearest productions. Several methods for
structural word matching have been reported in the literature\cite{Bunke}.
The basic algorithm states that the distance between two words is related to
the sequence of edit operations (substitution, insertion, and deletion)
required to transform one word into another. For any sequence of edit
operations a cost function $c\left( s\right) $ is considered 
\[
c\left( s\right) =\sum_{i=1}^{n}c\left( e_{i}\right) 
\]
$c\left( s\right) $ denotes the cost of a particular sequence $s$, and $%
c\left( e_{i}\right) $ the cost of a particular edit operation. The distance
between two words $x$ and $y$ is defined as the minimum cost of transforming
a word into another 
\[
d\left( x,y\right) =\min \left\{ c\left( s\right) |s\textnormal{ being a sequence
of edit operations transforming }x\textnormal{ into }y\right\} 
\]

When a word cannot be recognized because there is no production generating
some symbol in that word, a grammatical interpolation formula is used to
obtain the last terminal symbol $y_{p+1}$ in the right-hand side of the new
production 
\[
y_{p+1}=\textnormal{quantification}\left( \left( \sum_{i=1}^{n}d_{i}\right)
\sum_{i=1}^{n}\left( \frac{1}{d_{i}}y_{p+1}|_{\textnormal{prod }i}\right) \right) 
\]
This interpolation formula establishes a weighed average of the available
productions of the required order, on an average distance between similar
productions. The distance $d_{i}$ is the distance between the words $%
y_{1}...y_{p}U_{k}$ in the left-hand side of the existent productions and
the word of the new production.

Fig.9 shows the recognition results for the same grammar as considered in
Fig.8, when the grammatical interpolation procedure is applied. The effect
of the non-existent productions is reduced, and the number of symbols that
are not recognized much smaller. Grey arrows denote the interpolation of
non-existent productions and white arrows productions that could not be
obtained by interpolation.
\begin{figure}[tbh]
\begin{center}
\psfig{figure=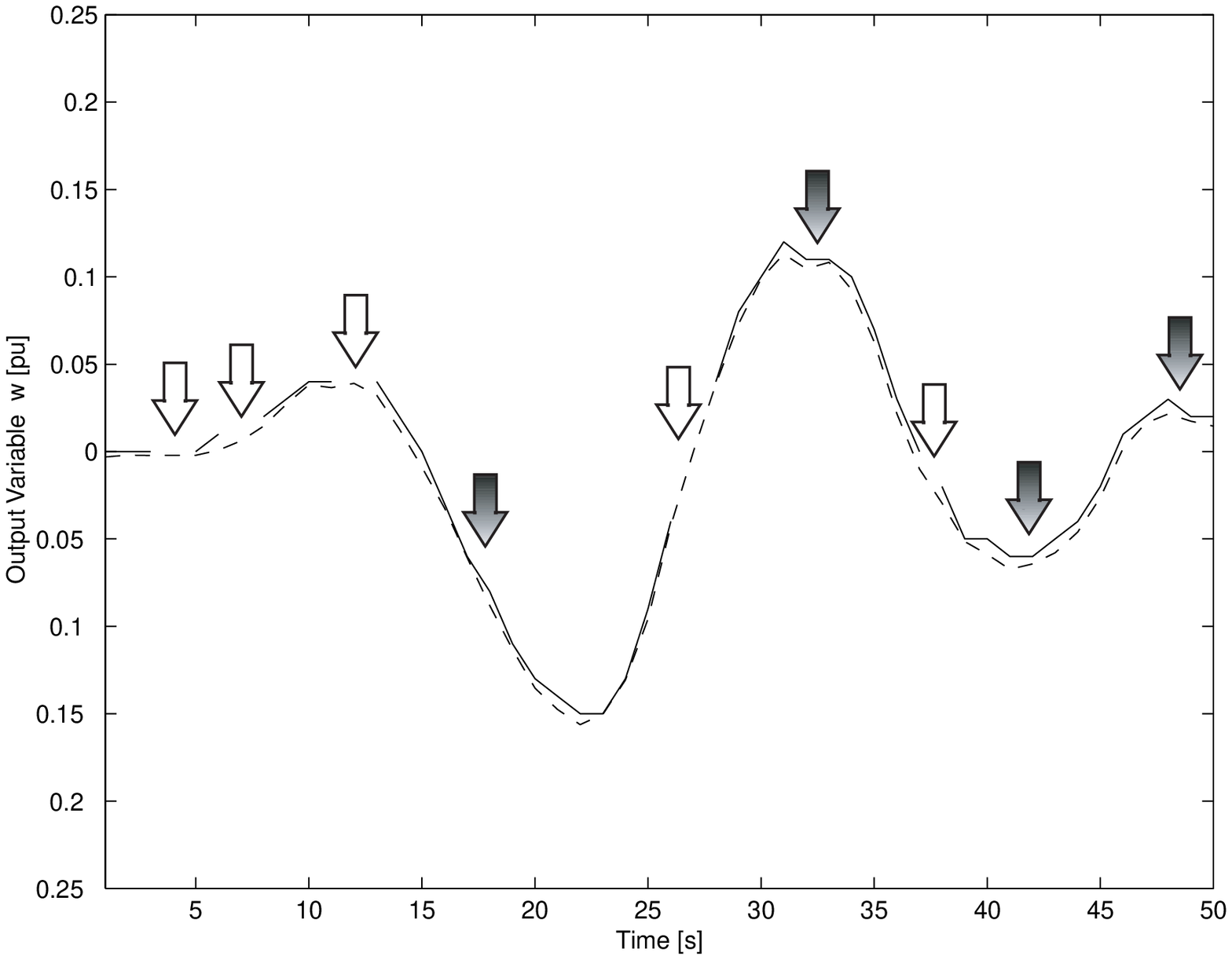,width=9truecm}
\end{center}
\caption[]{Formal language recognition with grammatical interpolation}
\end{figure}

By \textit{generalization} we mean the ability of an algorithms to perform
even under working conditions distinct of those observed during the learning
phase. The interpolation process as described above provides generalization
abilities to the algorithm. However this type of generalization is a local
effect in the working space. For example, consider the training set shown in
Fig.10.
\begin{figure}[tbh]
\begin{center}
\psfig{figure=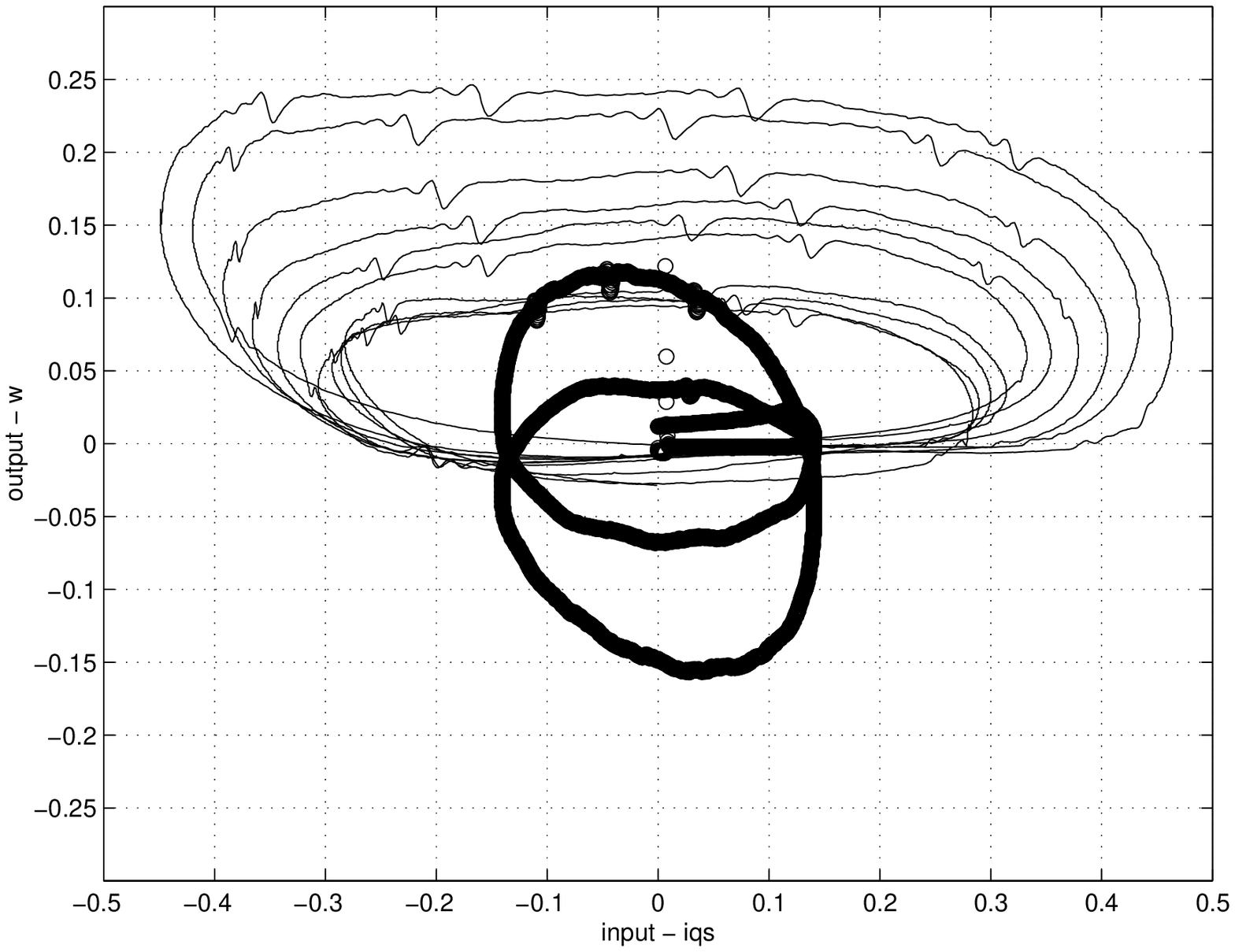,width=9truecm}
\end{center}
\caption[]{Restricted training and test data set for the electrical drive}
\end{figure}

In this case the training set contains only positive speed values. Fig.11
shows the modelling results that are obtained using this sample data.
Recognition fails for some of the domain areas not covered by the training
data set. Namely symbols that code for negative drive speed are not
recognized. Generalization has only a local effect. Therefore, experimental
training data must always cover a significant part of the working domain.
\begin{figure}[tbh]
\begin{center}
\psfig{figure=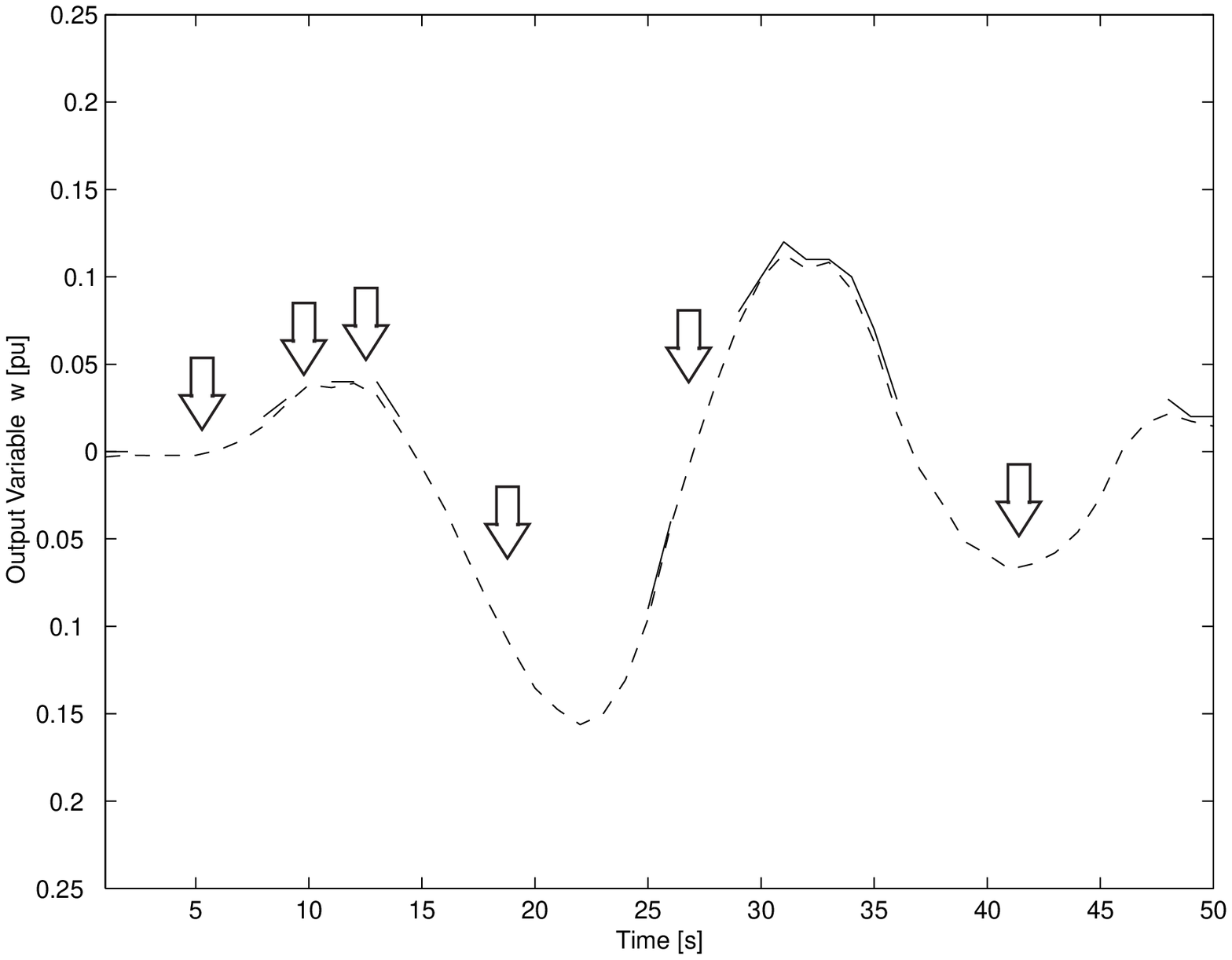,width=9truecm}
\end{center}
\caption[]{Generalization using grammatical interpolation}
\end{figure}

\section{Anomaly detection}

Any grammar $G$ codes for the class of patterns that belong to the language $%
L(G)$ the grammar generates. In this way one may use the grammar to
recognize well formed strings and reject anomalous ones. Once one has
learned the grammar of the strings generated by the system under normal
conditions, if at a later time there is some fault in the dynamic system,
strings that will be generated are not compatible with the learned grammar.
It is also reasonable to assume that this might occur already at an early
stage of the anomaly. The language algorithm might therefore be valuable as
a tool for early fault detection.
\begin{figure}[tbh]
\begin{center}
\psfig{figure=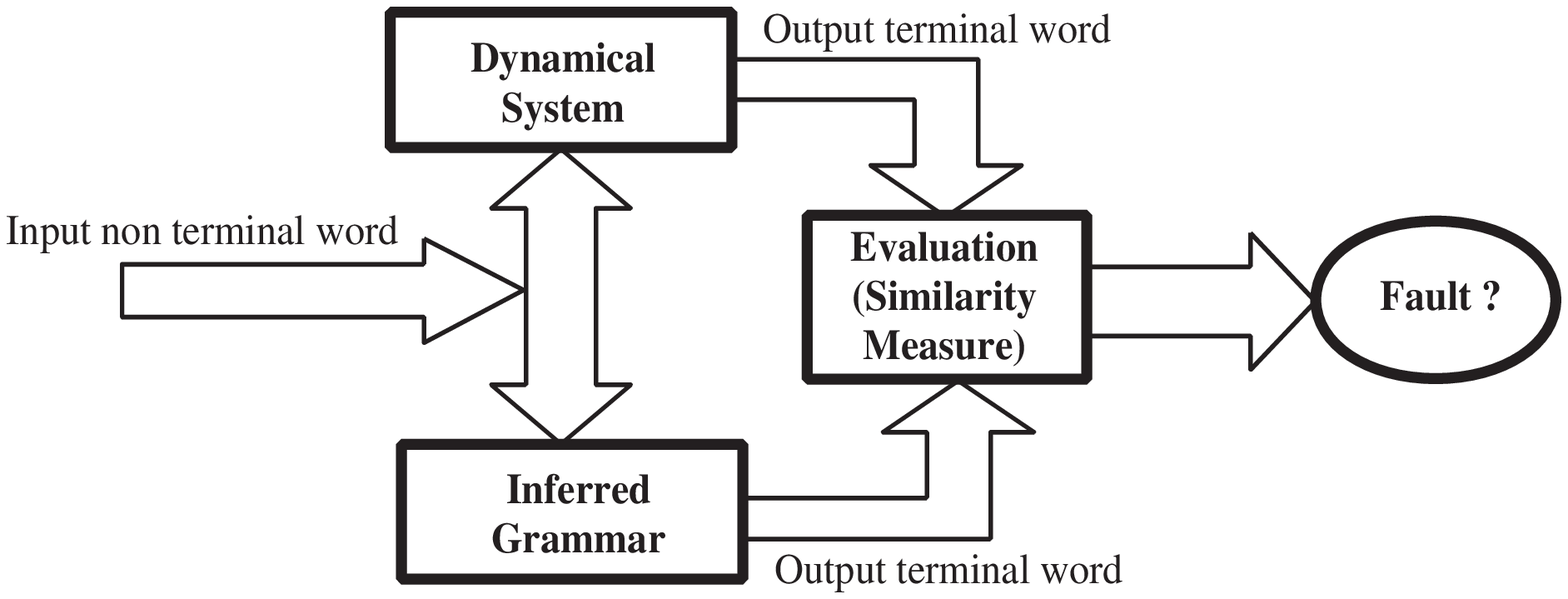,width=9truecm}
\end{center}
\caption[]{Fault detection}
\end{figure}

Detection of grammar anomalies is based on the same distance and similarity
measures as considered before, for grammar interpolation. The anomaly
detection algorithm (Fig.12) computes the distance between the string
generated by the dynamical system and the one generated by the learned
grammar. If this distance exceeds some threshold, a fault is reported.

As a first example, let us consider the dynamical system represented by Eq.(%
\ref{2.9}), with the same codification and inferred grammar as before. For
the anomaly detection test we consider a random input during 100 time steps.
A sustained anomaly was simulated between time steps 30 to 60, and a
spurious one between time steps 80 to 85. The anomaly simulation is simply
the replacement of the parameter 2 by 3 in Eq.(\ref{2.9}). The dynamical
system changes and therefore some of the strings that are generated do not
match the learned grammar. Whenever the distance between the
system-generated strings and those of the grammar exceeds a threshold, an
anomaly is reported.

Figs.13-15 display the anomaly detection results. In response to the control
variable both the learned grammar and the system produce terminal words.
These are shown in Fig.14, the output of the grammar as a continuous line
and the output of the dynamical system as a dashed line. They differ
whenever there is an anomaly. Fig.15 displays the distance between words
produced by the learned grammar and those of the dynamical system, for a
word length of 10 symbols. This distance exceeds the threshold only for the
sustained anomaly.
\begin{figure}[tbh]
\begin{center}
\psfig{figure=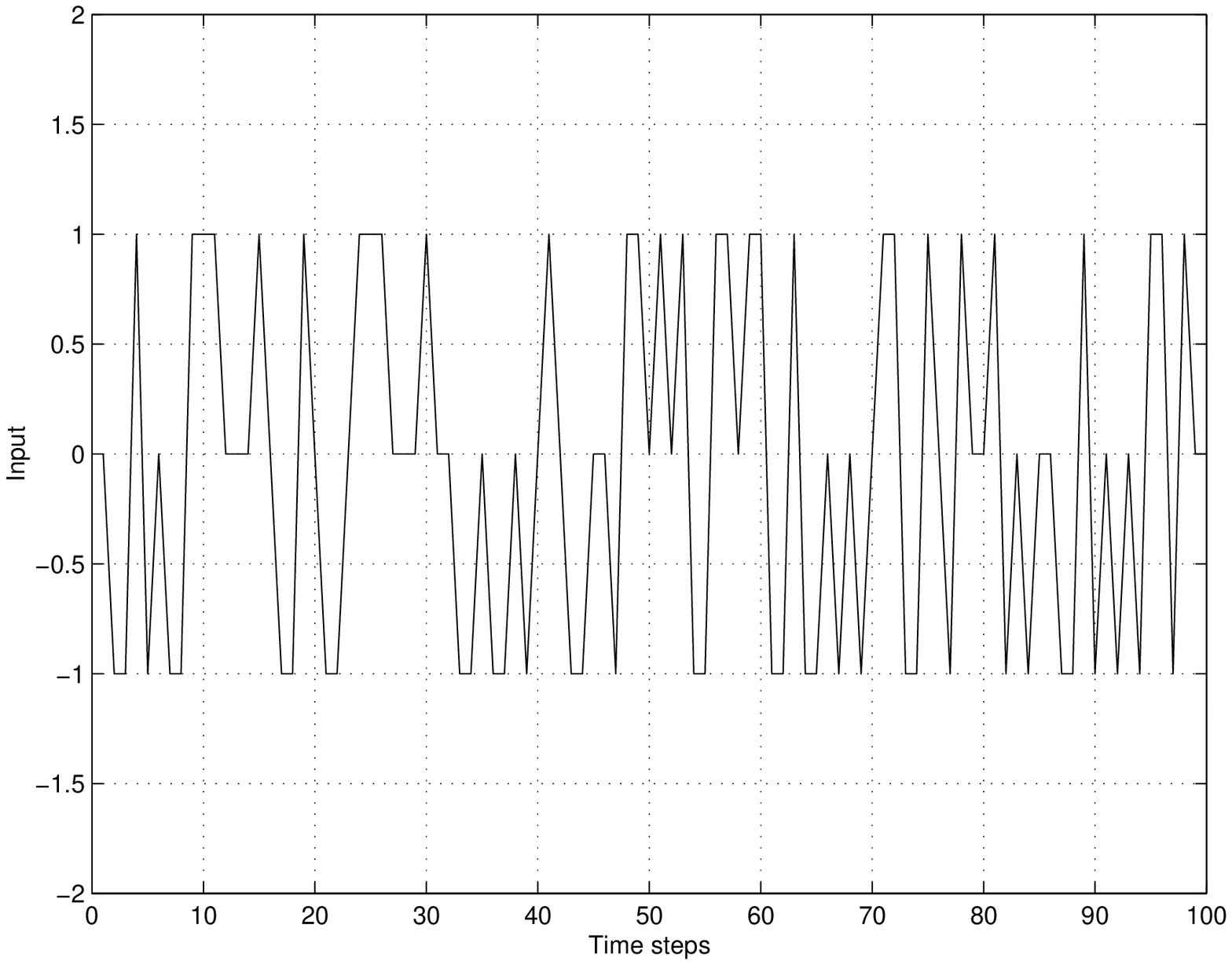,width=9truecm}
\end{center}
\caption[]{Control variable}
\end{figure}
\begin{figure}[tbh]
\begin{center}
\psfig{figure=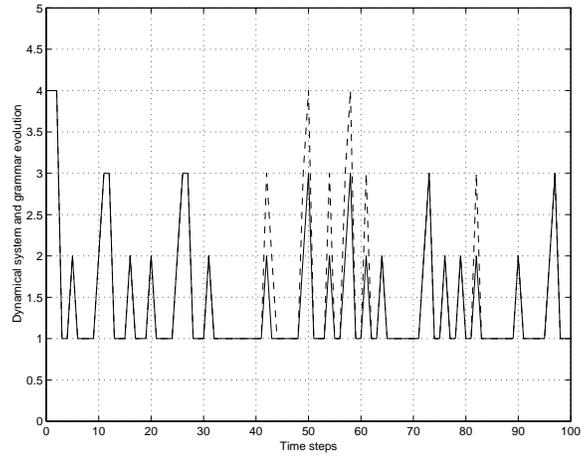,width=9truecm}
\end{center}
\caption[]{Learned grammar and actual system outputs
}
\end{figure}
\begin{figure}[tbh]
\begin{center}
\psfig{figure=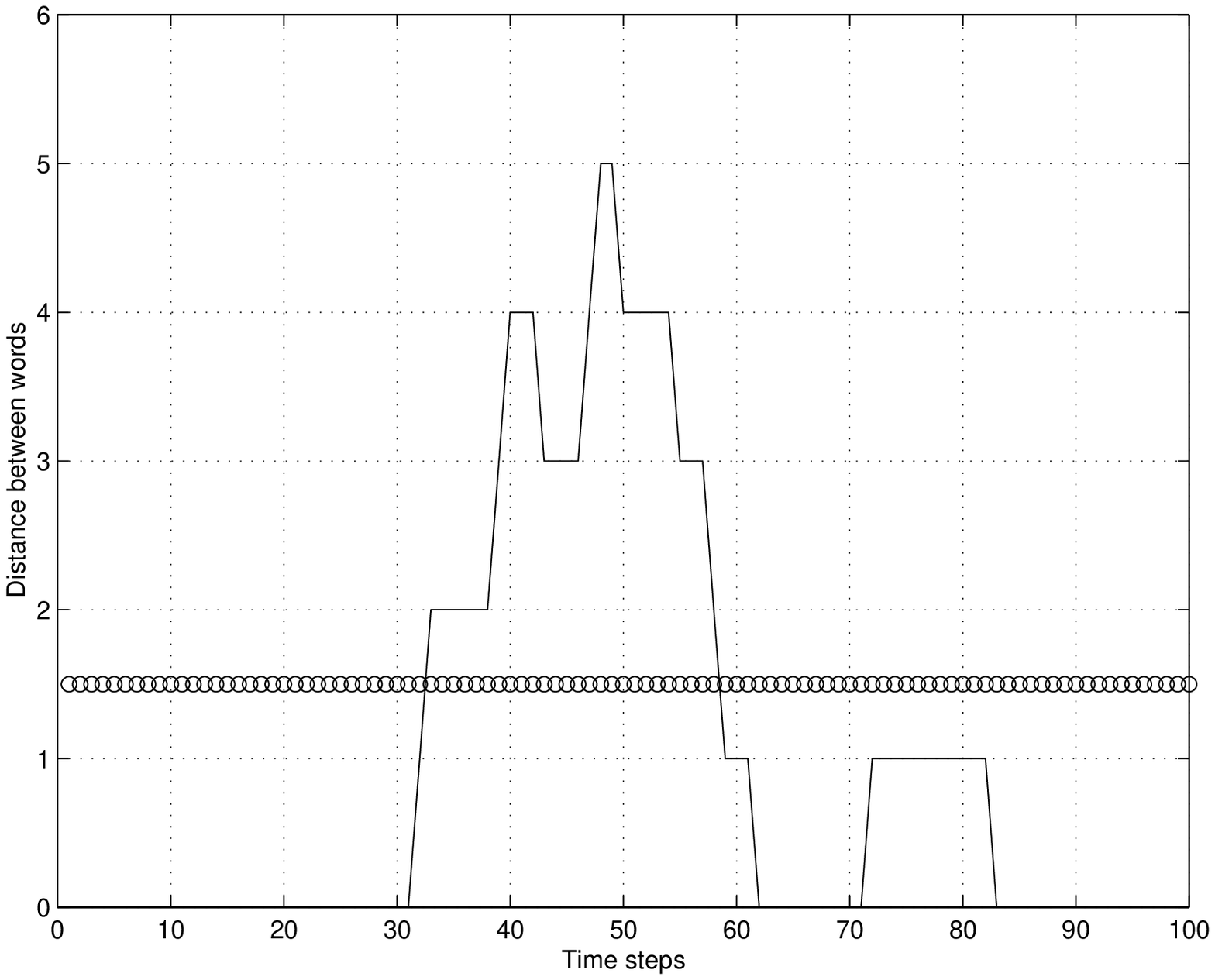,width=9truecm}
\end{center}
\caption[]{Distance between words}
\end{figure}

As a second example we consider anomaly detection in an induction motor. The
anomaly is a rotor broken bar, and the variable coded as a terminal alphabet
is the stator phase current. The learned grammar, representing the
experimental drive system, is generated with the motor operating without
anomaly. Fig16 shows the stator current deviation due to the presence of
the anomaly between time 1 and 2 (seconds). Fig.17 shows the distance
between the words generated by the learned grammar and those of the
induction motor. Soon after the onset of the anomaly, this distance begins
to exceed the threshold. A non-zero threshold must be defined because, as is
seen in the figure, some deviations are obtained even without anomalies due
to noise or even to small mismatches in the learned grammar.
\begin{figure}[tbh]
\begin{center}
\psfig{figure=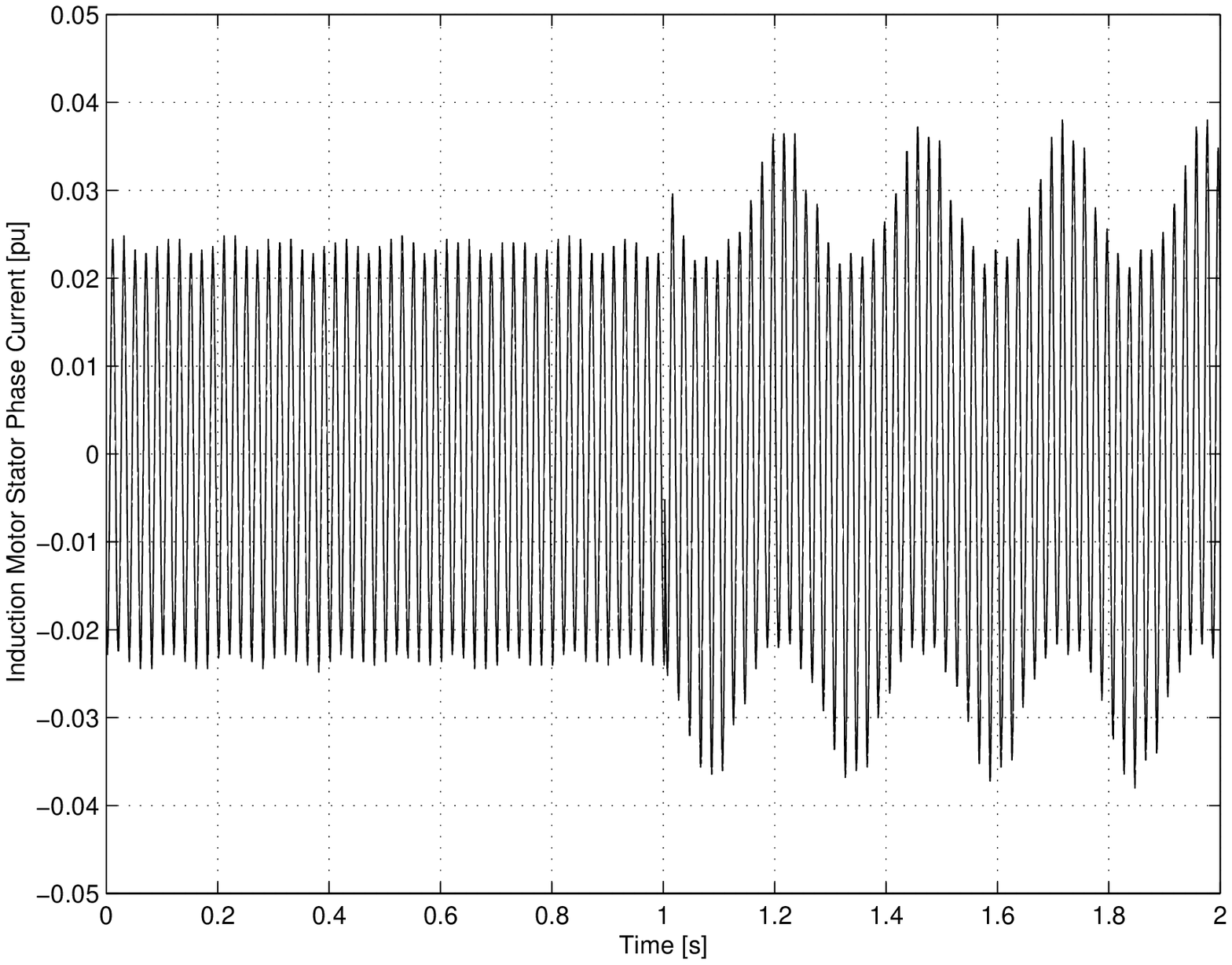,width=9truecm}
\end{center}
\caption[]{Induction motor stator phase current}
\end{figure}
\begin{figure}[tbh]
\begin{center}
\psfig{figure=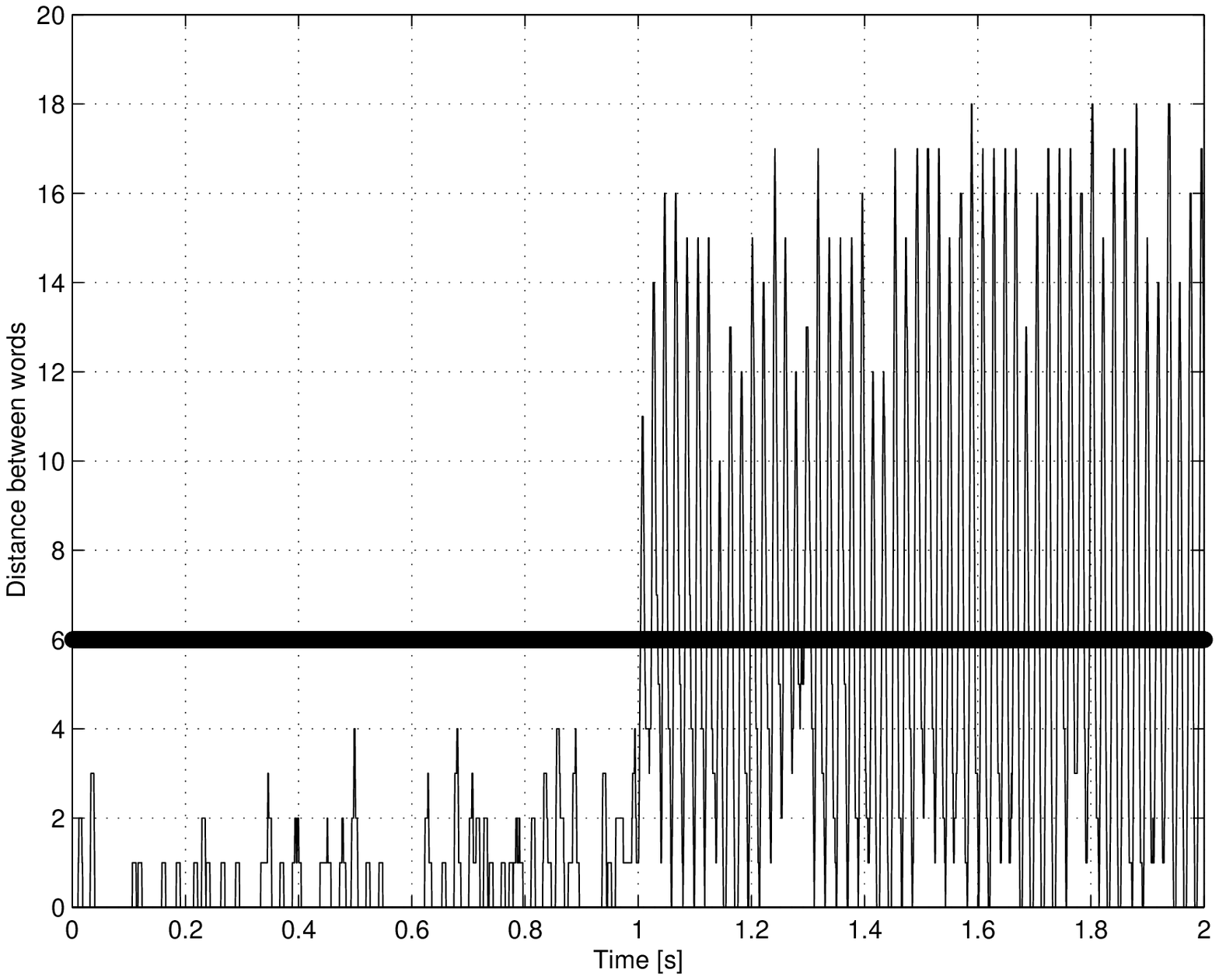,width=9truecm}
\end{center}
\caption[]{Distance between words}
\end{figure}

\section{Control}

Once the grammar generated by a dynamical system is learned, this grammar,
defining the structural features of the words produced by the linguistic
source, is a model for the source. So it is possible to use the productions
to predict the evolution of the system or to find the control string that
leads from one state to another.

The basic idea of the control algorithm is as follows. A string of terminal
symbols is given which represents the sequence of states that one wants the
system to follow. Knowing the initial state of the system, for each symbol
of the desired sequence of target states one chooses a production leading
the system from its present state to the desired one. At each step the
production that is chosen is the one that leads the system as close as
possible to the desired state. The procedure is repeated until that symbol
is reached and then one moves to the next target symbol.

As an example consider the dynamical system defined by Eq.(\ref{2.9}).
Suppose that the sequence of target states is '$cbcbaabcccbacc$'. Fig.18
shows the evolution of the system obtained by the control methodology
described above. The target sequence is shown as a dashed line and the
system evolution as a continuous line.
\begin{figure}[tbh]
\begin{center}
\psfig{figure=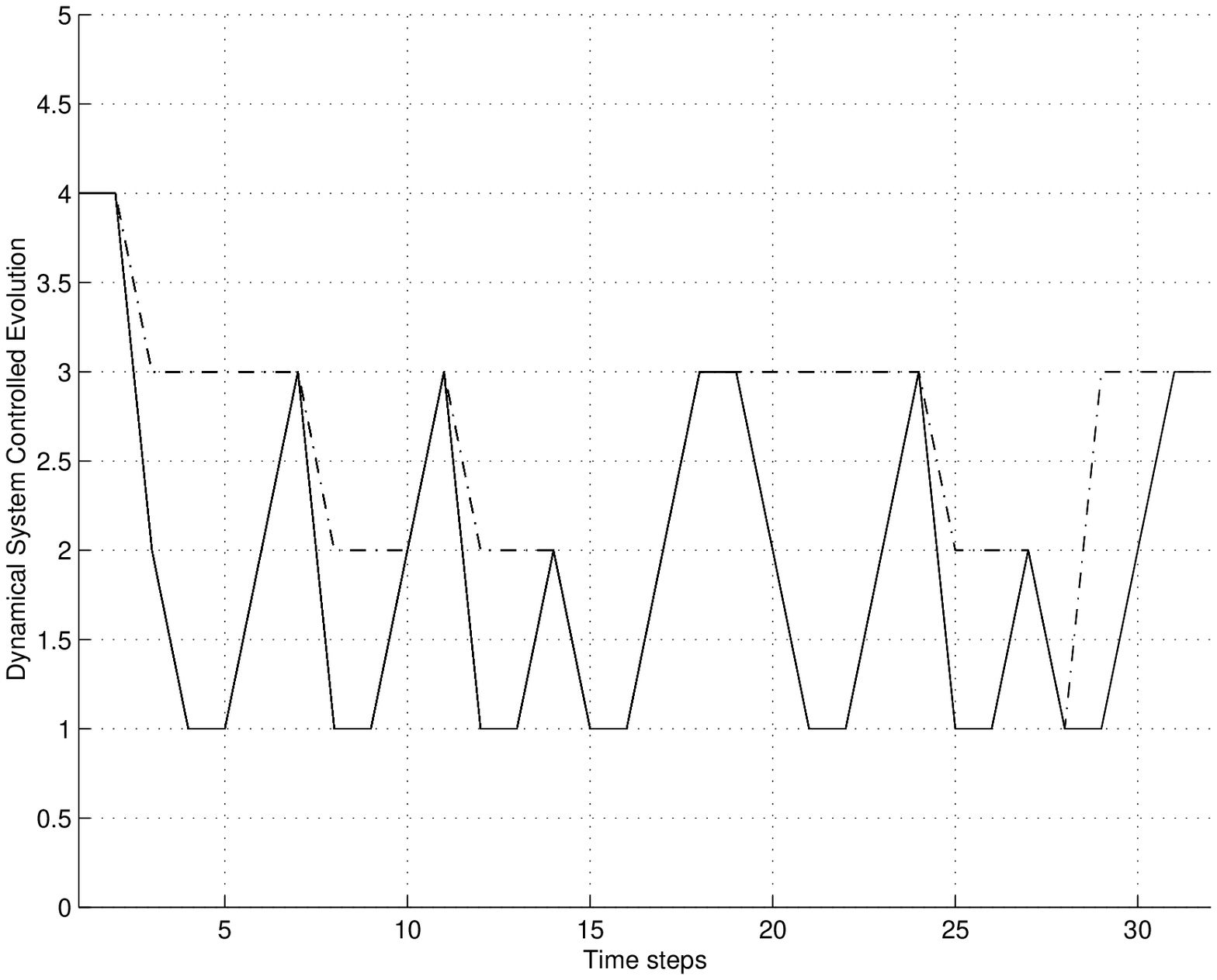,width=9truecm}
\end{center}
\caption[]{Controlled evolution of a dynamical system}
\end{figure}

Notice that to evolve from one target symbol to another the system may have
to pass through other terminal symbols. The control algorithm is robust in
the sense that whenever, at some time $t$, the expected evolution differs
from the one predicted by the learned grammar, due to some perturbation or
noise, it suffices to apply the same algorithm once more starting from the
state at time $t$.

\section{Conclusions and Remarks}

This paper proposes a characterization of controlled system dynamics by
formal language techniques. The applications we have been concerned with are
mostly electromechanical drives. However, the technique might also be useful
for many other dynamical systems or industrial processes.

An important issue is the choice of an appropriate quantification for the
feature space. This influences the choice and size of the alphabets and the
dimension and complexity of the grammars. Also, good coverage of the working
domain, in the learning stage, is essential to insure good generalization
properties.

Unlike other approaches, the nature of the rules that define the dynamical
system are not set up in advance, different types of productions being
established on-line according to the incoming words from the linguist
source. The method seems to provide good recognition results and a
reasonable quantitative accuracy. Fuzzy logics approaches are probably more
efficient when quantitative accuracy is very important and a good knowledge
already exists concerning the rules of the system. However, if the dynamics
is unknown or if pattern recognition is the important issue, the formal
language approach seems quite promising.

Fault detection in control systems, industrial plants and power networks is
a field where this work is being pursued and compared with immunity-based
learning systems.

\end{document}